\documentclass[10pt,twocolumn,letterpaper]{article}

\usepackage{iccv}              

\PassOptionsToPackage{table,dvipsnames}{xcolor}
\usepackage{xcolor}
\usepackage{colortbl}
\definecolor{iccvblue}{rgb}{0.21,0.49,0.74}
\usepackage[colorlinks=true, citecolor=iccvblue, linkcolor=red, urlcolor=red]{hyperref}
\usepackage{makecell}
\usepackage{multirow}
\usepackage{booktabs}
\usepackage{fontawesome5}
\usepackage{pifont}
\usepackage{stix}

\definecolor{lit-org}{rgb}{0.95, 0.72, 0.35}
\definecolor{lit-red}{rgb}{1, 0.449, 0.449}
\definecolor{lit-blue}{rgb}{0.449, 0.727, 0.88}

\newcommand{\J}{$\mathcal{J}$\xspace}
\newcommand{\F}{$\mathcal{F}$\xspace}
\newcommand{\JF}{$\mathcal{J}\&\mathcal{F}$\xspace}

\newcommand{\Fo}{$\dot{\mathcal{F}}$\xspace}
\newcommand{\JFo}{$\mathcal{J}\&\dot{\mathcal{F}}$\xspace}
\newcommand{\JFnd}{$\mathcal{J}\&\dot{\mathcal{F}}_d$\xspace}
\newcommand{\JFnr}{$\mathcal{J}\&\dot{\mathcal{F}}_r$\xspace}

\def\paperID{*****} 
\def\confName{ICCV}
\def\confYear{2025}

\title{The 1st Solution for MOSEv2 Challenge 2025: \\ 
Long-term and Concept-aware Video Segmentation via SeC}

\author{
    Mingqi Gao\textsuperscript{1}\quad
    Jingkun Chen\textsuperscript{2}\quad
    Yunqi Miao\textsuperscript{3}\quad
    Gengshen Wu\textsuperscript{4}\quad
    Zhijin Qin\textsuperscript{1}\quad
    Jungong Han\textsuperscript{1}\footnotemark[1] \\
 \textsuperscript{1}Tsinghua University \;
  \textsuperscript{2}University of Oxford \;
  \textsuperscript{3}University of Warwick \;
  \textsuperscript{4}City University of Macau \\
  {\tt\scriptsize $\{$im.mingqi,yunqimiao709$\}$@gmail.com,  jingkun.chen@eng.ox.ac.uk, gswu@cityu.edu.mo, $\{$qinzhijin,jghan$\}$@tsinghua.edu.cn}
}

\begin{document}
\maketitle
\renewcommand{\thefootnote}{\fnsymbol{footnote}}
\footnotetext[1]{Corresponding author.}
\begin{abstract}

This technical report explores the MOSEv2 track of the LSVOS Challenge, which targets complex semi-supervised video object segmentation. By analysing and adapting SeC\footnotemark[5], an enhanced SAM-2 framework, we conduct a detailed study of its long‑term memory and concept‑aware memory, showing that long‑term memory preserves temporal continuity under occlusion and reappearance, while concept‑aware memory supplies semantic priors that suppress distractors; together, these traits directly benefit several MOSEv2’s core challenges. Our solution achieves a \JFo score of 39.89\% on the test set, ranking 1st in the MOSEv2 track of the LSVOS Challenge. 

\end{abstract}    
\section{Introduction}
\label{sec:intro}

Large-scale Video Object Segmentation (LSVOS)\footnotemark[4] workshop focuses on segmenting objects in complex and realistic video scenarios. It provides large-scale training and testing datasets, a unified evaluation platform, challenges and discussions. This year, a new track MOSEv2~\cite{ding2025mosev2} (Complex VOS) was introduced. It brings additional challenges related to object disappearance and reappearance, occlusion, visibility conditions, object size and type, and the need for knowledge-guided reasoning. Together with existing tracks, Classic VOS (MOSEv1~\cite{ding2023mose} and LVOS~\cite{hong2023lvos}) and Referring VOS (MeViS 2023~\cite{ding2023mevis}), this year’s LSVOS builds a competitive platform encouraging robust and practical solutions for real-world applications. 

In parallel with LSVOS, the organising team has released the MeViS 2024 dataset~\cite{ding2025mevis}, which extends MeViS 2023 from language‑only expressions to multimodal guidance integrating audio and language, and introduces a novel Referring Motion Expression Generation (RMEG) task. These encourage the development of practical VOS applications and provide the community with a richer, high‑quality, and challenging benchmark and evaluation platform, and further motivate the community to pursue holistic, object‑level perception, understanding, and generation architectures. 

This report focuses on the MOSEv2 Challenge, targeting Complex VOS under the semi-supervised setting~\cite{gao2023deep}. Similar to classic VOS, the task provides pixel-level annotations of the target in the first frame, and the goal is to segment the same object or region across the rest of the video. Due to its ability to separate arbitrary target regions from background at the pixel level, semi-supervised VOS is widely used in downstream tasks such as video editing~\cite{wang2025videodirector}, dense video understanding~\cite{tang2025caption}, and embodied AI~\cite{chen2025m}.

\footnotetext[4]{\url{https://lsvos.github.io/}}
\footnotetext[5]{\url{https://github.com/OpenIXCLab/SeC}}

To push research closer to realistic scenarios, the MOSE series~\cite{ding2023mose,ding2025mosev2} introduces videos, high-quality annotations, diverse categories, and complex challenges. Unlike datasets such as DAVIS~\cite{perazzi2016benchmark} and YouTube-VOS~\cite{xu2018youtube}, where methods saturate (around 90\%), the MOSE series focuses on more difficult conditions, including heavy occlusion, similar objects, and crowded distractors. MOSEv2 goes even further than MOSEv1 by adding more videos, denser annotations, and increased difficulty through stronger occlusions, more frequent disappearance and reappearance, more distractors, and more small targets. It also includes diverse lighting conditions (e.g., nighttime or underwater), weather (e.g., rain, snow, or fog), and non-physical targets (e.g., shadows), making the dataset more realistic. New evaluation metrics are also proposed to better analyse these challenges. As a result, many existing methods drop to around 50\% performance, showing limitations in object perception, long-term modelling, and temporal consistency. 

Existing semi-supervised VOS relies on memory-based designs~\cite{oh2019video}, where past segmentation results are stored to segment future frames. Early work focused on datasets with handcrafted designs, which limit generalisation to real scenes. Recently, large models like SAM-2~\cite{ravi2024sam2}, trained on tens of millions of annotated masks, gained popularity due to strong generalisation, flexible interactions, and simple architecture. As base model, SAM-2 uses a first-in-first-out strategy to manage memory and achieves strong performance with minimal tuning. Later methods aim to better manage historical information, such as SAMURAI~\cite{yang2024samurai} (motion-guided memory selection), SAM2Long~\cite{ding2024sam2long} (tree-based memory), and DAM4SAM~\cite{videnovic2025distractor} (distractor-aware memory). However, these methods are all built on training-free SAM-2 and still inherit its limitations in modelling temporal context, which makes them less effective on MOSEv2’s core challenges like occlusion, disappearance, and concept-driven scenes.

Our solution uses SeC~\cite{zhang2025sec}, a SAM-2 framework with enhanced concept modelling. Experiments on MOSEv2 show SeC’s properties fit its core challenges well: heavy occlusion, frequent disappearance and reappearance, and distractors. On MOSEv2, our \JFo is 53.34\% on the validation set and 39.89\% on the test set.

\section{Related Work}
\label{sec:relate}

\paragraph{Semi-supervised Video Object Segmentation.}

Existing methods are based on memory, first introduced in STM~\cite{oh2019video}. They selectively store intermediate segmentation results during inference, allowing the model to consider diverse object appearances and adapt as the video progresses over time. By establishing fine-grained correspondence between intermediate frames and the current frame, associated segmentation masks can be propagated to the current frame. This simple, efficient design greatly improves video object segmentation performance and remains widely used today.

Before SAM-2, the community focused on small-scale, domain-specific datasets and models. The main advances came from cross-frame correspondence metrics, as seen in STCN~\cite{cheng2021rethinking} and XMem~\cite{cheng2022xmem}, or from object-level constraints on correspondence, such as in Cutie~\cite{cheng2024putting}. Although these methods achieved high-quality results on in-domain datasets, their generalisation to real-world data remains limited. 

With tens of millions of training data and a large model, SAM-2~\cite{ravi2024sam2} improves generalised VOS. Later work focused on organising and selecting high-quality memories. SAMURAI~\cite{yang2024samurai} predicts motion and occlusion from temporal cues. It adaptively selects memory masks or features, removing noisy memory, to reduce error propagation and guide tracking in next frame. SAM2Long~\cite{ding2024sam2long} replaces the greedy single path memory with a constrained tree search strategy. It keeps a number of branches. At each frame, it generates multiple candidates by uncertainty, accumulates scores, and keeps high scoring paths. The final frame selects the best path, curbing cascading memory errors. DAM4SAM~\cite{videnovic2025distractor} scores and filters memories by target and distractor similarity and confidence. It removes drift prone memories and keeps memories for recovery, improving tracking under occlusion and interference. 

Although these methods perform well in challenging scenes, they are all training-free extensions based on SAM-2. Therefore, their performance is still limited by the lack of long-term context and strong object-level constraints. As a result, they cannot effectively handle typical challenges in MOSEv2, such as severe occlusion, frequent disappearance and reappearance, and knowledge-dependent scenes. 

\vspace{-1em}
\paragraph{Combining SAM-2 with Large Vision-Language Models (LVLMs).}

LVLMs~\cite{liu2023visual,chen2024expanding,Qwen2.5-VL} are widely used in language-guided video object segmentation (also called Referring VOS). This is because they align vision and text, use strong training data, and benefit from the world knowledge and language interface provided by large language models. Given a video and a text description of the target, LVLMs generate tokens related to the target on key frames. With the help of SAM-2, which supports prompts and accurate object perception, these tokens are used as prompts to segment the key frames. The remaining frames are segmented by tracking with SAM-2. Thanks to the reasoning ability and world knowledge of LVLMs, Referring VOS has moved towards more complex tasks, such as VISA~\cite{yan2024visa}, VideoLISA~\cite{bai2024one}, and Sa2VA~\cite{sa2va}. Since both LVLMs and Referring VOS rely on vision-language interaction, using LVLMs for Referring VOS is natural and has shown strong results.

Besides Referring VOS, LVLMs also have great potential for semi-supervised VOS. Their high-level understanding of video content helps filter out many distractors with a similar appearance. The recently proposed SeC~\cite{zhang2025sec} is the first to explore this. It uses InternVL-2.5~\cite{chen2024expanding} as LVLMs, building a ``concept prior'' across frames that acts as semantic memory, instead of fine-grained memory in earlier methods. During inference, SeC balances concept reasoning and feature matching, and adjusts computation based on scene transitions. This helps maintain robust segmentation under large appearance changes and scene transitions. 
\section{Solution}
\label{sec:method}

\begin{figure*}[!ht]
\centering
\includegraphics[width=\linewidth]{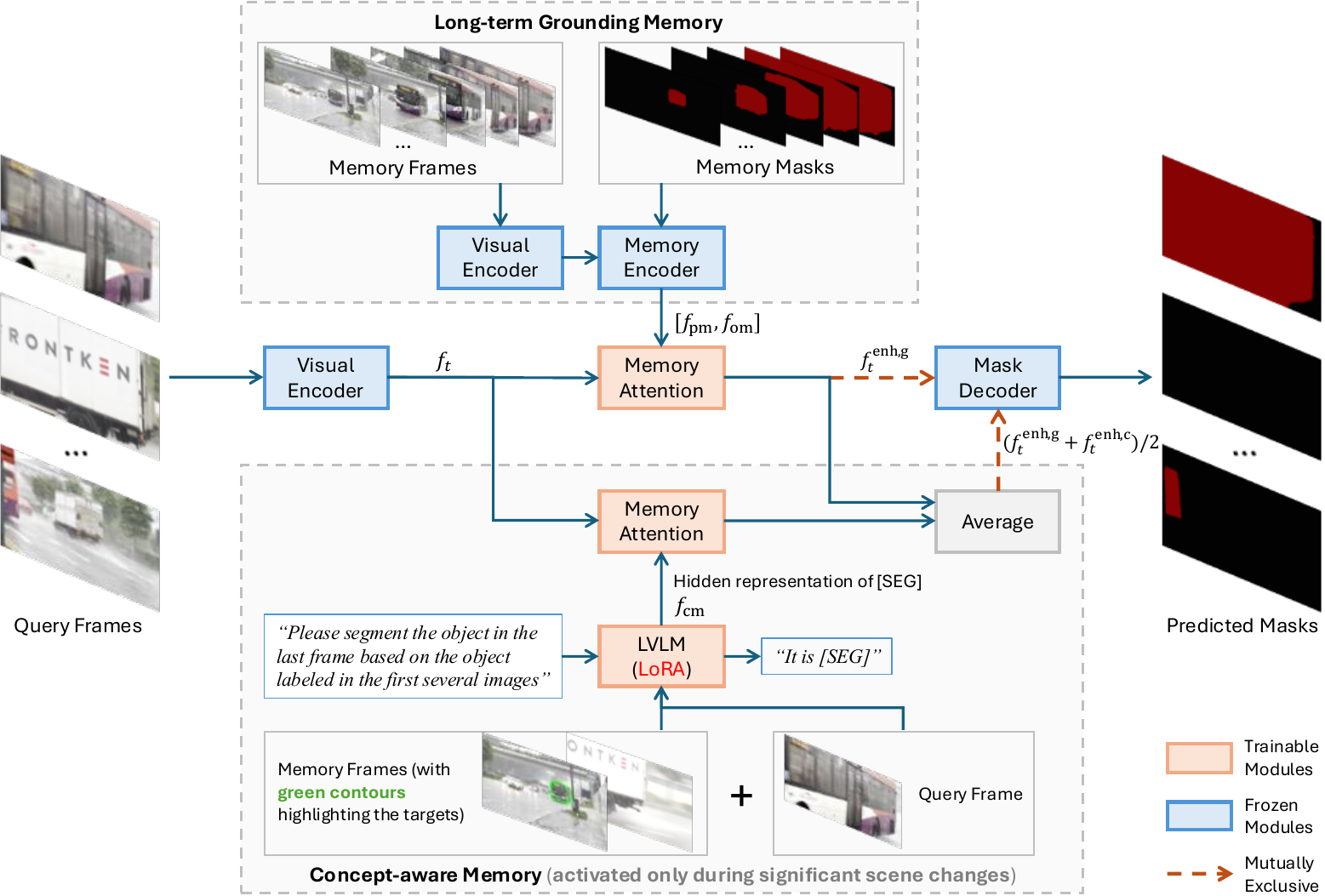}
   \caption{Technical details of SeC~\cite{zhang2025sec}. The architecture includes two main modules. (1) Long-term Grounding Memory uses the long-range and fine-grained correspondence between memory frames (up to 22 frames) and query frames to transfer targets' labels to the query frames, enabling accurate localisation and segmentation. (2) Concept-aware Memory uses a vision-language model (InternVL-2.5~\cite{chen2024expanding}) to analyse the semantics and context of targets in the memory frames. The [SEG] token's hidden representation gathers high-level information about the targets. Both types of memory enhance the query features to inject target information and improve robustness from different aspects. For efficiency, Concept-aware Memory is only activated when significant scene changes are detected. }
\label{fig:diagram}
\end{figure*}

Our solution uses SeC~\cite{zhang2025sec}, a SAM-2-based framework with enhanced concept modelling. For more details about model training and data, please refer to the original paper. As shown in Fig.~\ref{fig:diagram}, we provide more technical details of SeC, which is built based on InternVL-2.5-4B~\cite{chen2024expanding}
and SAM-2-Large~\cite{ravi2024sam2}. The former builds semantic-level memory for the targets, while the latter focuses on fine-grained cross-frame matching and strong object perception. Given an input video with $T$ frames ($\mathcal{V}=\{v_t\in \mathbb{R}^{H\times W\times 3}\}^{T}_{t=1}$) and the first-frame annotation $m_1$, SeC generates pixel-level masks of the target object for the remaining frames $\mathcal{M}=\{m_t\in \mathbb{R}^{H\times W}\}_{t=2}^T$. The inference of each frame selectively goes through two types of memory:

\vspace{-1em}
\paragraph{Long-term Grounding Memory.}
Following SAM-2, it has two parts. For example, at frame $t$, the memory includes: 1) Pixel Memory: $f_\mathrm{pm}\in \mathbb{R}^{N_l\times C\times h\times w}$, built from the first frame and frames from ($t-N_l+1$) to ($t-1$). $C$, $h$, $w$ denote the number of channels and spatial size ($h=H/16$, $w=W/16$). $N_l$ is the memory size. It encodes both dense pixel features and the predicted masks of those frames. 2) Object Memory: $f_\mathrm{om}\in \mathbb{R}^{N_l\times C}$, also built from the first frame and frames from ($t-N_l+1$) to ($t-1$). It uses intermediate features from the mask prediction process, which implicitly represent object-level information. The Pixel Memory is first flattened, then concatenated with the Object Memory to form the final memory. Given the features of the $t^\mathrm{th}$ frame $f_t\in \mathbb{R}^{C\times h\times w}$, it is enhanced with the memory via 4 layers of self-attention and cross-attention, achieving $f_t^\mathrm{enh,g}=\mathrm{Cross\_Attn}(q=\mathrm{Self\_Attn}(f_t),kv=[f_\mathrm{pm},f_\mathrm{om}])$. These features are then decoded for the segmentation result. 

Unlike SAM-2, SeC allows a much larger memory size ($N_l=22$), with 24 frames during training. The memory attention modules are adjusted accordingly. Compared to SAM-2 (memory size is 7 during training/inference), SeC is clearly better at capturing long-term cross-frame relations, which helps in complex spatiotemporal scenarios.

\vspace{-1em}
\paragraph{Concept-aware Memory.}
It is built from a set of memory frames with masks. Unlike long-term memory, it keeps at most $N_c=7$ video frames, maintained in a FIFO manner.

This step is not always active. It only runs when a major scene change is detected. For detecting changes, SeC uses the Bhattacharyya distance between HSV histograms. In the challenge, we use a threshold of 0.35 to decide whether to apply the concept-aware memory. The features for mask decoding become the mean of $f_t^\mathrm{enh,c}$ and $f_t^\mathrm{enh,g}$ when the concept memory is activated. 

\section{Experiments}
\label{sec:label}

\begin{figure*}[!t]
\centering
\includegraphics[width=\linewidth]{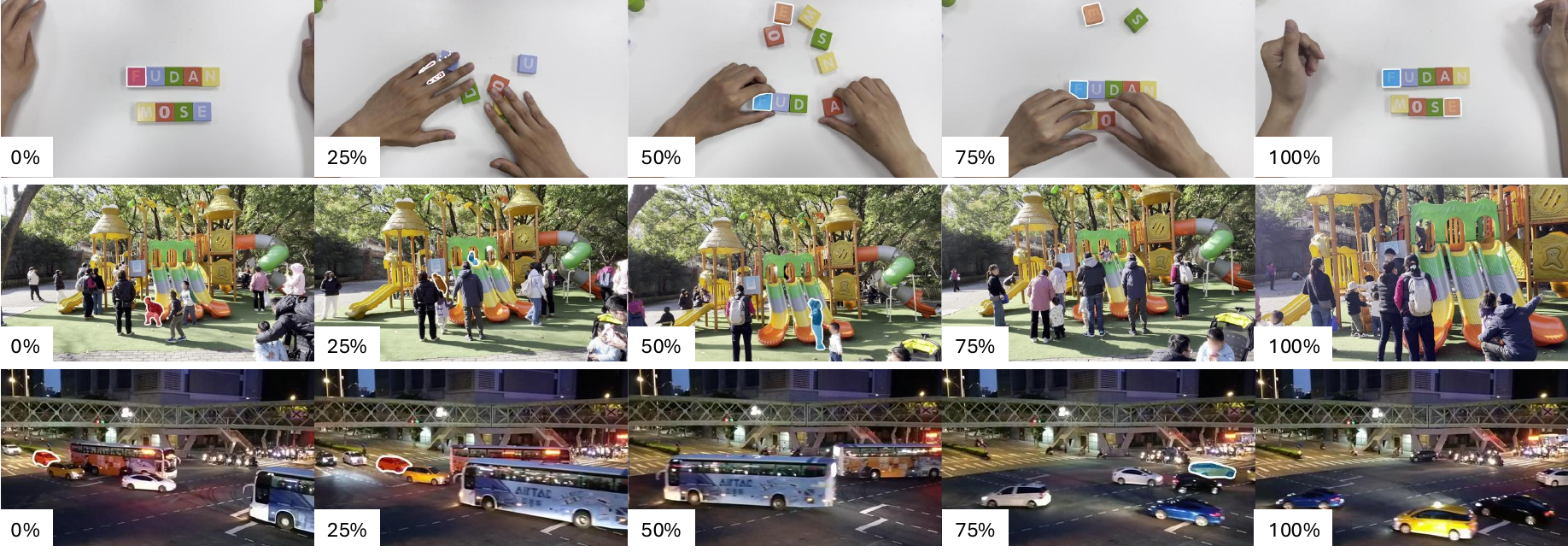}
\vspace{-1.5em}
   \caption{Qualitative ablations on the long-term memory on the MOSEv2 validation videos (from top to bottom: 0t51bkn3, 2wzfma2s, and xhvp0o0a). {\footnotesize\textbf{\textcolor{lit-blue}{\faSquare}}} 
   \textbf{\textcolor{lit-blue}{Blue}} and {\footnotesize\textbf{\textcolor{lit-org}{\faSquare}}} \textbf{\textcolor{lit-org}{Orange}} masks come from the SeC with $N_l=22$ and $N_l=7$. {\footnotesize\textbf{\textcolor{lit-red}{\faSquare}}} \textbf{\textcolor{lit-red}{Red}} masks are their overlap. The percentage indicates the position of the corresponding frame in the video. For the three videos above, the \JFo scores of blue masks outperform those of orange masks by 59.69, 32.23, and 40.58, respectively. }
\label{fig:abl1}
\end{figure*}

\begin{figure*}[!t]
\centering
\includegraphics[width=\linewidth]{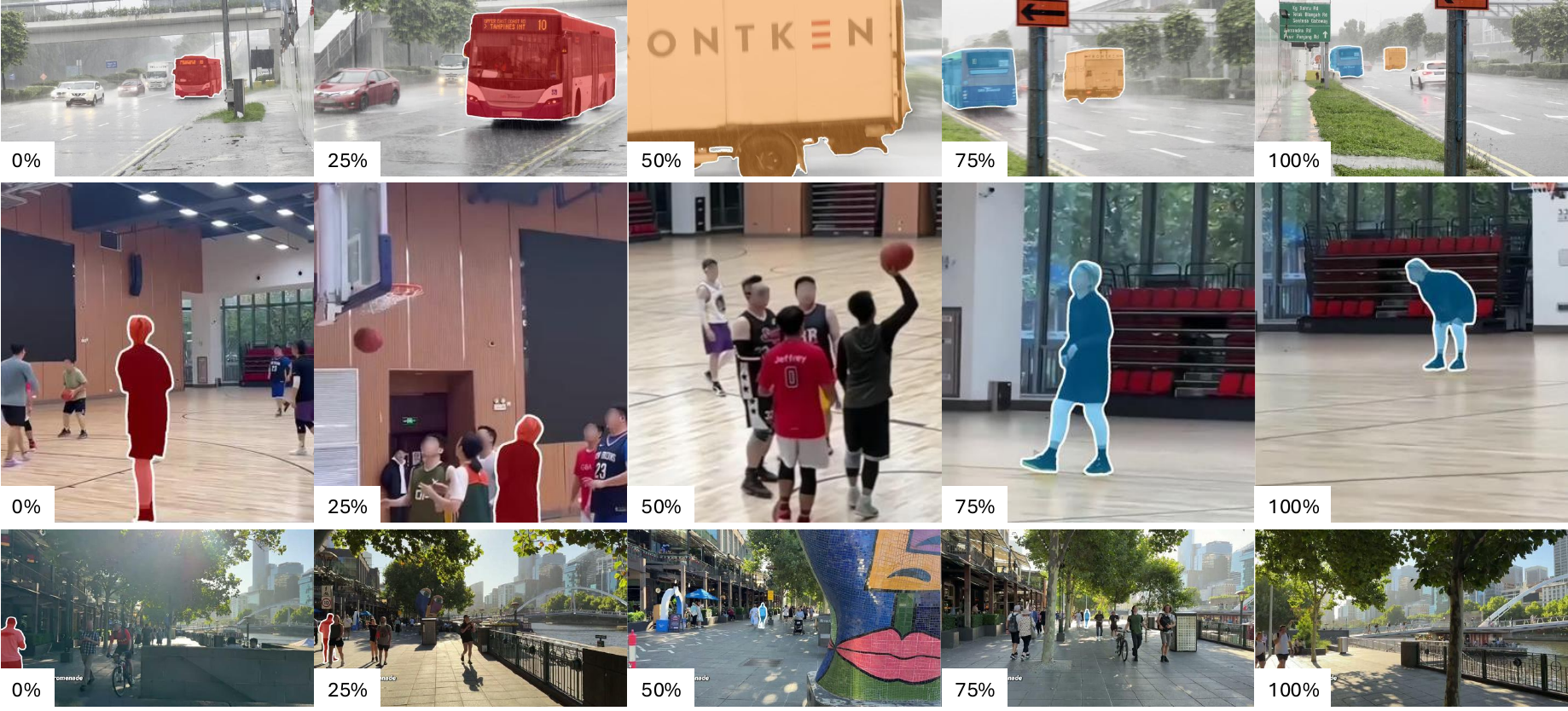}
\vspace{-1.5em}
   \caption{Qualitative ablations on the concept-aware memory on the MOSEv2 validation videos (from top to bottom: 9mv2g80y, jtefx29o, and y38n4pox). {\footnotesize\textbf{\textcolor{lit-blue}{\faSquare}}} 
   \textbf{\textcolor{lit-blue}{Blue}} and {\footnotesize\textbf{\textcolor{lit-org}{\faSquare}}} \textbf{\textcolor{lit-org}{Orange}} masks come from the SeC with (threshold=0.35) and without the concept-aware memory. {\footnotesize\textbf{\textcolor{lit-red}{\faSquare}}} \textbf{\textcolor{lit-red}{Red}} masks are their overlap. The percentage indicates the position of the corresponding frame in the video. For the three videos above, the \JFo scores of blue masks outperform those of orange masks by 55.82, 28.95, and 36.47, respectively. }
\label{fig:abl2}
\end{figure*}

This section first presents the quantitative results on the MOSEv2 test set. We then provide several ablations to investigate the impact of different SeC configurations on the MOSEv2 validation set, to derive insights for future research.

Unlike traditional metrics that only include \JF, \J, and \F, MOSEv2 introduces three new metrics to better reflect its challenges. (1) \Fo measures adaptive contour accuracy and is more suitable for evaluating small objects. (2) \JFnd focuses on accuracy during the target disappearance period. (3) \JFnr evaluates performance after the target reappears. These metrics allow for more accurate and targeted evaluation under the complex conditions in MOSEv2. 

\subsection{Main Results}

Tab.~\ref{tab:test} shows our quantitative results on MOSEv2 test set. 

\begin{table}\renewcommand{\arraystretch}{1.3}
\tabcolsep=0.1cm
\footnotesize
  \centering
  \begin{tabular}{p{0.18\columnwidth}|p{0.09\columnwidth}<{\centering}p{0.09\columnwidth}<{\centering}p{0.09\columnwidth}<{\centering}p{0.09\columnwidth}<{\centering}p{0.09\columnwidth}<{\centering}p{0.09\columnwidth}<{\centering}p{0.09\columnwidth}<{\centering}}
    \toprule
    Participant & \JFo & \J & \Fo & \JFnd & \JFnr & \F & \JF \\
    \midrule
    \rowcolor{blue!17!white}
    mmm & \textbf{39.89} & \textbf{39.02} & \textbf{40.76} & 57.15 & \textbf{19.00} & \textbf{42.35} & \textbf{40.68} \\
    qqqqaaaa & \underline{39.70} & \underline{38.87} & \underline{40.53} & 57.84 & \underline{18.67} & \underline{42.09} & \underline{40.48} \\
    limjduni & 37.87 & 36.99 & 38.75 & \textbf{64.68} & 12.05 & 40.06 & 38.52 \\
    waaaaaaaaaa & 35.77 & 34.98 & 36.56 & \underline{61.95} & 11.82 & 37.90 & 36.44 \\
    springggg & 35.39 & 34.63 & 36.15 & 61.89 & 11.60 & 37.47 & 36.05 \\
    \rowcolor{gray!10}
    \multicolumn{8}{c}{$\mathbf{\cdots}$} \\
    \bottomrule
  \end{tabular}
  \caption{Quantitative comparisons on the MOSEv2 test set. Our scores are \colorbox{blue!17!white}{highlighted} in the blue row. \textbf{Bold} and \underline{underlined} entries denote the best and second-best results.}
  \label{tab:test}
\end{table}

\subsection{Ablations}

\paragraph{Long-term Memory.} 

Tab.~\ref{tab:ablmem} evaluates how the size of long-term grounding memory affects segmentation. More memory frames yield clear gains, suggesting future work could further explore long‑term spatiotemporal interaction. Qualitative results in Fig.~\ref{fig:abl1} show that a larger memory size also helps localise occluded targets. A likely reason is that including more frames increases the chance that the memory stores more diverse and higher‑quality target states, enabling more discriminative decisions during memory attention and improving robustness to distractors. Long-term interaction can also enhance target localisation on the query frame. 

\begin{table}\renewcommand{\arraystretch}{1.3}
\tabcolsep=0.1cm
\footnotesize
  \centering
  \begin{tabular}{p{0.17\columnwidth}|p{0.09\columnwidth}<{\centering}p{0.09\columnwidth}<{\centering}p{0.09\columnwidth}<{\centering}p{0.09\columnwidth}<{\centering}p{0.09\columnwidth}<{\centering}p{0.09\columnwidth}<{\centering}p{0.09\columnwidth}<{\centering}}
    \toprule
    \scriptsize{Memory Size} & \JFo & \J & \Fo & \JFnd & \JFnr & \F & \JF \\
    \midrule
    7 & 51.07 & 49.33 & 52.82 & \textbf{71.82} & 28.93 & 55.56 & 52.44 \\
    \rowcolor{blue!17!white}
    22 & \textbf{53.34} & \textbf{51.47} & \textbf{55.20} & 70.10 & \textbf{33.74} & \textbf{57.93} & \textbf{54.70} \\
    \bottomrule
  \end{tabular}
  \caption{Ablations on the memory size ($N_l$). $N_l=7$ is the default setting in SAM-2. Our setting during the challenge is \colorbox{blue!17!white}{highlighted}. }
  \label{tab:ablmem}
\end{table}

\vspace{-1em}
\paragraph{Concept-aware Memory.}

Tab.~\ref{tab:ablconcept} shows the impact of the scene-change threshold on segmentation results. The scores show concept-aware memory brings improvements. The qualitative results in Fig.~\ref{fig:abl2} illustrate its advantages, especially in keeping target localisation after occlusion and staying robust against distractors. For example, in the first case, the target is a ``bus''. After it disappears, without concept-aware memory, the segmentation drifts to a similar-category ``truck''. Even when the bus reappears, the mistake cannot be fixed. Across all cases in Fig.~\ref{fig:abl1}, concept-aware memory shows the ability to understand differences between similar objects and apply this understanding during segmentation. This helps reduce errors caused by over-reliance on fine-grained correspondence. We also found that using concept-aware memory on all frames leads to a noticeable performance drop. This opens space for future work on using the concept module more effectively and combining it better. 

\begin{table}\renewcommand{\arraystretch}{1.3}
\tabcolsep=0.1cm
\footnotesize
  \centering
  \begin{tabular}{p{0.17\columnwidth}|p{0.09\columnwidth}<{\centering}p{0.09\columnwidth}<{\centering}p{0.09\columnwidth}<{\centering}p{0.09\columnwidth}<{\centering}p{0.09\columnwidth}<{\centering}p{0.09\columnwidth}<{\centering}p{0.09\columnwidth}<{\centering}}
    \toprule
    Threshold & \JFo & \J & \Fo & \JFnd & \JFnr & \F & \JF \\
    \midrule
    0 (all w/ C)* & 52.49 & 50.63 & 54.35 & 69.43 & 32.83 & 57.08 & 53.86 \\
    \rowcolor{blue!17!white}
    0.35 & \underline{53.34} & 51.47 & \underline{55.20} & \underline{70.10} & \underline{33.74} & \underline{57.93} & \underline{54.70} \\
    0.5 & \textbf{53.44} & \textbf{51.58} & \textbf{55.30} & \textbf{70.16} & \textbf{33.92} & \textbf{58.03} & \textbf{54.81} \\
    0.7 & \underline{53.34} & \underline{51.48} & 55.19 & 69.98 & 33.61 & 57.90 & 54.69 \\
    1 (w/o C) & 53.29 & 51.44 & 55.14 & 69.98 & 33.51 & 57.85 & 54.64 \\
    \bottomrule
  \end{tabular}
  \caption{Ablations on the scene-change threshold. 0 indicates using concept (C) on ALL frames, and 1 indicates that concept (C) is not considered. Our setting during the challenge is \colorbox{blue!17!white}{highlighted}. *Note that for the 0 (all w/ C) setting, we did not run the full video set. Enabling the LVLM on all frames increased the inference time by around 20 times. Therefore, we only ran this setting on 50 videos, and copied the other results from our default setting.}
  \label{tab:ablconcept}
\end{table}

\vspace{-0.5em}
\section{Conclusion}
\label{sec:conclusion}

Our study demonstrates that the design of SeC, combining long-term memory with concept-aware reasoning, naturally aligns with several key challenges of MOSEv2. Through targeted analysis and configuration, we show that SeC can effectively address occlusion, disappearance, and distractors, achieving strong results.
{
    \small
    \bibliographystyle{ieeenat_fullname}
    \bibliography{main}
}

\end{document}